\documentclass{article}

\PassOptionsToPackage{numbers,compress}{natbib}
\usepackage[preprint]{neurips_2026}

\usepackage[utf8]{inputenc}
\usepackage[T1]{fontenc}
\usepackage{hyperref}
\usepackage{url}
\usepackage{graphicx}
\usepackage{booktabs}
\usepackage{amsfonts}
\usepackage{amsmath}
\usepackage{multirow}
\usepackage[table]{xcolor}
\usepackage{wrapfig}
\usepackage{subcaption}
\usepackage{nicefrac}
\usepackage{microtype}
\usepackage[accsupp]{axessibility}
\graphicspath{{./}{../}}

\title{AsySplat: Efficient Asymmetric 3D Gaussian Splatting for Long-Sequence Scene Modeling} 

\author{%
\textbf{Yingji Zhong$^{1}$ \quad
Dave Zhenyu Chen$^{2}$ \quad
Fuzhao Ou$^{3}$ \quad
Youyu Chen$^{2}$} \\
\textbf{Zhihao Li$^{2}$ \quad
Lanqing Hong$^{2}$ \quad
Dan Xu$^{1}$} \\
$^{1}$HKUST \quad
$^{2}$Huawei Noah's Ark Lab \quad
$^{3}$CityU
}

\begin{document}

\maketitle
\vspace{-3mm}
\begin{abstract}
  Recent generalizable 3D Gaussian Splatting models have advanced long-sequence novel view synthesis (NVS), but at the cost of substantial redundant computation. We identify that the redundancy can be mitigated based on two observations: (i) high-precision geometry is not strictly required for high-quality NVS; (ii) appearance learning is generally easier than geometry recovery. 
  Motivated by these insights, we propose an asymmetric architecture that decouples geometry and appearance modeling. The geometry branch processes coarse-grained tokens with most of the parameters for multi-view reconstruction, while the appearance branch operates on fine-grained tokens to capture details using significantly fewer parameters. The two branches interact through bilateral connections, enabling mutual guidance for their respective tasks. 
  This task-aware asymmetry reduces the computational redundancy and allocates the computation more judiciously, thereby increasing parameter efficiency and enabling smaller models to achieve strong performance. 
  On 32-view 960P inputs, our model matches optimization-based methods while delivering nearly 800$\times$ speedup, and surpasses the zero-shot performance of state-of-the-art generalizable models with markedly fewer parameters and reduced training/inference overhead, achieving an overall efficiency improvement. 
  The project page is at \href{https://zhongyingji.github.io/asysplat/}{https://zhongyingji.github.io/asysplat/}. 
\end{abstract}\vspace{-3mm}

\vspace{-3mm}
\section{Introduction}
\label{sec:intro}

3D Gaussian Splatting (3DGS)~\cite{kerbl20233d,yu2024mip,mallick2024taming} has emerged as an effective representation for novel view synthesis (NVS). Recent research has shifted toward generalizable 3DGS models~\cite{charatan2024pixelsplat,chen2024mvsplat,chen2024lara,tang2024lgm,xu2024grm,zhang2025gs,xu2025depthsplat}, also referred to as Gaussian Reconstruction Models, which predict 3D Gaussians from networks, bypassing per-scene optimization through data-driven priors. While most methods focus on sparse inputs, real-world captures often consist of long sequences, making such a setting more practical. With the rise of high-resolution devices and the growing expectations of visual detail, it becomes increasingly important to extend generalizable 3DGS methods to handle high-resolution, long-sequence inputs~\cite{ziwen2025long}.

\begin{figure}[t]
  \centering
  \includegraphics[width=0.99\linewidth]{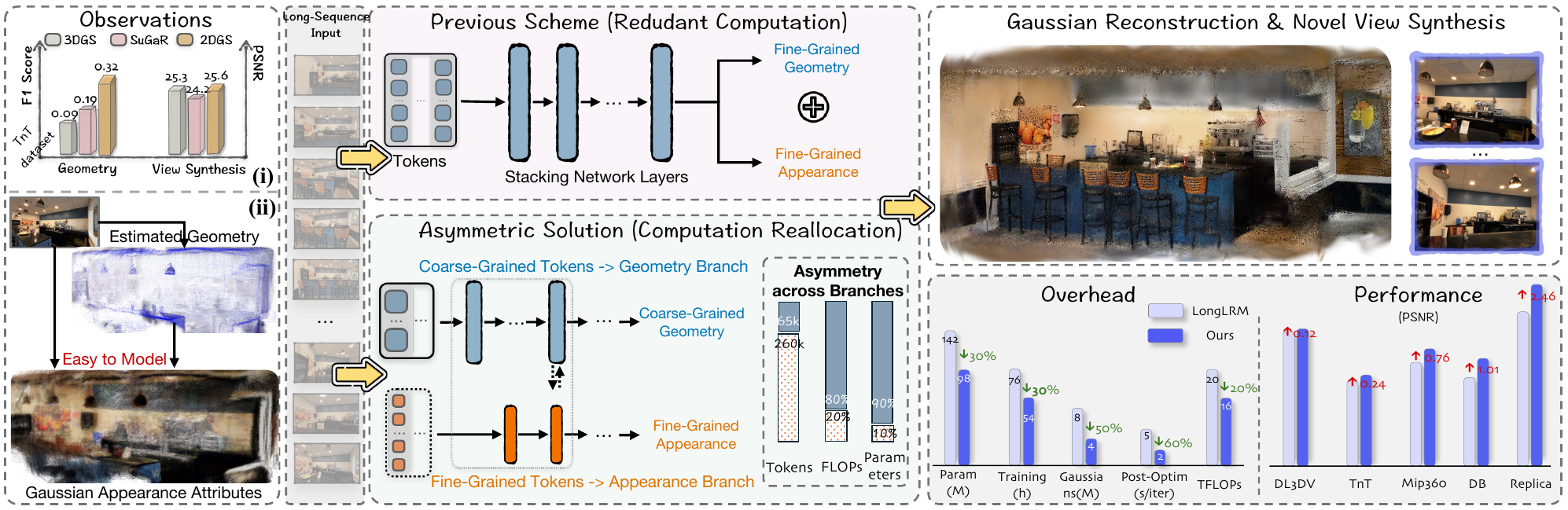}\vspace{-2mm}
  \caption{Our AsySplat reduces computational redundancy in the long-sequence Gaussian reconstruction model, by decoupling it into asymmetric geometry and appearance branches. The asymmetry spans token granularity, computation allocation, and parameter distribution, which is motivated by two observations: \textbf{(i)} High-precision geometry is not a must for high-quality rendering. \textbf{(ii)} Once geometry is estimated, high-precision appearance attributes are easy to model by assigning pixel-aligned RGB evidence to Gaussians. This design enables a markedly smaller model to achieve strong performance with reduced training and inference overhead. }
  \label{fig:teaser}\vspace{-4mm}
\end{figure}

Most current methods like LongLRM~\cite{ziwen2025long} stack numerous layers~\cite{vaswani2017attention,gu2024mamba} to predict depth and Gaussian appearance attributes directly from patchified tokens. This design concentrates all computation on inter-frame token interactions. This costly process stems from the need for small patch sizes to achieve high-quality rendering. This raises a fundamental question: Does such simplistic stacking introduce computational redundancy that drives up parameter demand? In this work, we explore whether a more deliberate allocation of computation can improve parameter efficiency in NVS, thereby empowering compact models to achieve competitive performance.

In this work, we highlight two observations that are crucial for computation allocation but overlooked in prior generalizable 3DGS methods: 
\textbf{(i)} \emph{High-precision geometry is not a must for high-quality rendering. }While geometry matters, the high-fidelity 3DGS rendering largely arises from the alpha compositing of Gaussians during splatting, rather than from highly precise geometry. This is consistent with reports of high rendering quality despite modest performance in surface reconstruction~\cite{guedon2024sugar,huang20242d}, as illustrated in Fig.~\ref{fig:teaser}. 
\textbf{(ii)} \emph{High-precision appearance attributes are necessary, while they are generally easier to model than geometry recovery. }Once multi-view depth is established, mapping RGB images to Gaussian appearance attributes is relatively straightforward, as it mainly involves assigning pixel-aligned RGB evidence to Gaussians. This property is also explored in scene generation works~\cite{yu2025wonderworld}. 
These insights prompt us to reconsider prevailing designs that allocate identical computation agnostically to geometry and appearance modeling~\cite{zhang2025gs,ziwen2025long}. 

Motivated by these insights, we propose an \emph{Asymmetric Gaussian Splatting} (AsySplat) model that reallocates computation by decoupling the modeling of geometry and appearance into two branches, and aligning each with its precision needs and learning difficulty. This decoupled design is asymmetric, manifested in input granularity, computation, and parameter allocation.  
\textbf{(i)} \emph{Geometry branch} processes coarse-grained tokens patchified with large patch sizes, reflecting the observation that high-precision geometry is not a must for NVS. Since multi-view reconstruction is intrinsically harder---requiring exhaustive inter-frame correspondence for depth estimation, we allocate the majority of parameters to this branch to ensure sufficient capacity for geometry recovery. 
With fewer tokens, the cost of computationally intensive inter-frame token interactions is greatly reduced. 
\textbf{(ii)} \emph{Appearance branch} models the Gaussian appearance attributes from fine-grained tokens patchified with small patch sizes. Fine-grained tokens are critical for preserving details and thus necessary for high-fidelity NVS. Since appearance prediction is relatively easier, we use slimmer feature channels and far fewer parameters than in the geometry branch (roughly 10$\%$ of the total). 
The two branches communicate via bilateral connections: the appearance branch injects fine details into the geometry branch, while the geometry branch supplies depth cues to guide the prediction of Gaussian attributes. The asymmetric structure resolves a key dilemma of prior works, which inevitably drag fine-grained tokens into computationally expensive inter-frame token interactions. 
In contrast, we retain fine-grained tokens for detailed rendering while keeping them out of those heavy interactions.
AsySplat not only provides a practical recipe for removing computation redundancy by allocating the computation more judiciously, but also improves per-parameter efficiency and enables smaller models to achieve strong performance. 

We adopt a hybrid structure~\cite{hatamizadeh2025mambavision,ziwen2025long} in the geometry branch to control the training overhead. However, excessive reliance on Mamba~\cite{gu2024mamba} degrades performance~\cite{ziwen2025long}, since state-space models are less effective for the explicit matching required by multi-view depth estimation.
We therefore reinvest the computation saved by our asymmetric design into additional attention~\cite{vaswani2017attention} layers to capture contextual cues more effectively, which is crucial for compact models. However, incorporating more attention layers increases training time. To mitigate this overhead, we introduce a sparse attention module that partitions each token’s context based on a 2D grid structure. This carefully designed module not only substantially reduces training time, e.g., a 40$\%$ time reduction in the computationally heaviest stage, but also maintains competitive performance.

Our contributions are summarized as follows: \vspace{-1.mm}
\begin{itemize}
    \item We introduce an asymmetric design to tackle computational redundancy in generalizable 3DGS for high-resolution and long-sequence inputs. This asymmetry enables a more judicious allocation of computation and parameters, leading to high parameter efficiency.
    \item We design a sparse attention module that allows us to retain more attention layers for higher performance, while not incurring large training overhead. 
    \item Experiments show that given 32-view 960P inputs, AsySplat achieves performance comparable to optimization-based methods with a speedup of 800$\times$. It surpasses prior generalizable 3DGS in zero-shot performance across three benchmarks, reducing parameters by 30$\%$, training time by 30$\%$, inference FLOPs by 20$\%$, Gaussians by 50$\%$, and achieving 60$\%$ faster post-optimization.

\end{itemize}

\section{Related Work}
\label{sec:related}
\vspace{-1mm}
\noindent \textbf{3D Gaussian Splatting for Novel View Synthesis. }In recent years, 3D Gaussian Splatting (3DGS)~\cite{kerbl20233d,yu2024mip} has emerged as a highly effective scene representation method for novel view synthesis (NVS). While Neural Radiance Fields (NeRFs) are capable of producing high-quality renderings~\cite{mildenhall2020nerf,barron2021mip,barron2022mip,barron2023zip}, 3DGS surpasses them not only in rendering fidelity but also in optimization efficiency and real-time rendering performance. 
Similar to the development of NeRF~\cite{muller2022instant,chen2022tensorf,zhong2024cvt,zhong2025empowering,zhenxing2022switch,pumarola2021d,azinovic2022neural}, subsequent 3DGS researches have extended in various directions, including reducing optimization time~\cite{mallick2024taming,chen2025dashgaussian}, handling sparse input views~\cite{zhu2025fsgs,zhang2024cor,zhong2025taming,chen2025quantifying}, large-scale scene reconstruction~\cite{lin2024vastgaussian,kerbl2024hierarchical,wang2025hyrf}, dynamic scene modeling~\cite{wu20244d,gao2024gaussianflow}, pose-free optimization~\cite{fu2024colmap}, and improving geometry for mesh recovery~\cite{chen2024pgsr,wolf2024gs2mesh,huang20242d}.

\noindent \textbf{Generalizable 3D Gaussian Splatting. }To circumvent the time-consuming per-scene optimization required by conventional 3DGS, recent research has explored generalizable approaches that directly predict scene representations from multi-view images in a feed-forward manner. Unlike generalizable NeRF methods, which typically output a 3D volume~\cite{yu2021pixelnerf,wang2021ibrnet,chen2021mvsnerf} or a triplane~\cite{hong2023lrm} for point querying during ray marching, generalizable 3DGS often employs pixel-aligned unprojection~\cite{charatan2024pixelsplat,szymanowicz2024splatter}. This technique maps Gaussian attributes predicted on 2D image planes into 3D space using estimated depth. To enhance multi-view depth estimation, several strategies have been introduced, such as leveraging epipolar geometry~\cite{charatan2024pixelsplat}, cost volume construction~\cite{chen2024mvsplat,chen2024mvsplat360}, or incorporating pre-trained depth models~\cite{xu2025depthsplat}. Some studies~\cite{zhang2025gs} further eliminate all 3D inductive biases, allowing the model to learn geometric constraints entirely from data.
Following the progress of pose-dependent generalizable 3DGS, a growing line of research explores pose-free generalizable 3DGS~\cite{ye2024no,xu2025freesplatter,kang2025selfsplat,jiang2025anysplat}.

However, all the aforementioned methods focus on sparse views, typically using no more than 16 views.  
Recent works~\cite{ziwen2025long,kang2026ilrm} extend generalizable 3DGS to high-resolution, long-sequence inputs, modeling scenes with up to 32 views at 960P resolution under known camera poses. We adopt this setting due to its practical importance for real-world applications.  
LongLRM~\cite{ziwen2025long} employs a stacking architecture and points out that the vast number of tokens patchified from high-resolution, long-sequence inputs leads to substantial computational overhead during both training and inference, mainly due to intensive token interactions. To alleviate this, LongLRM integrates the state-space model Mamba~\cite{gu2024mamba} to handle long-range dependencies more effectively. In this work, we advance beyond LongLRM by identifying computational redundancy within token interactions. We introduce an asymmetric architecture that mitigates this redundancy through computation reallocation, guided by the precision requirements and learning difficulty of geometric and appearance modeling for NVS. The more judicious allocation increases parameter efficiency and enables us to use smaller models to achieve strong performance. 

\begin{figure}[t]
  \centering
  \includegraphics[width=0.99\linewidth]{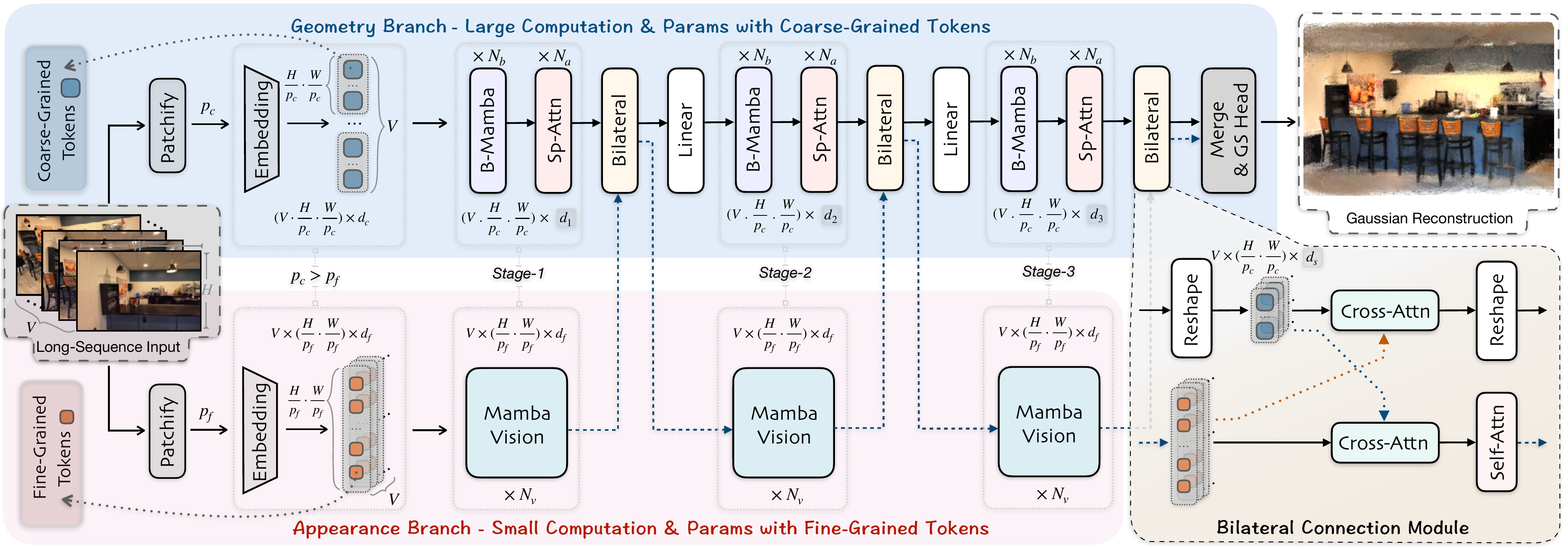}\vspace{-2mm}
  \caption{\textbf{Framework overview of AsySplat. }
  For Gaussian reconstruction from long-sequence inputs, AsySplat reallocates computation by decoupling the modeling of geometry and appearance into two branches, aligning each with its precision needs and learning difficulty. 
  The geometry branch processes coarse-grained tokens through three hierarchical stages with progressively increasing network capacity, leveraging bidirectional Mamba~\cite{zhu2024vision} and a sparse attention module (Fig.~\ref{fig:method_sp}) to model the complete \emph{inter-frame} token interactions. The appearance branch handles fine-grained tokens using lightweight MambaVision~\cite{hatamizadeh2025mambavision} with token interactions restricted to \emph{intra-frame}. 
  A bilateral connection module enables cross-branch interaction, and the final outputs from both branches are merged for Gaussian prediction. 
  The asymmetry across branches is reflected in token granularities, computation, and parameter distribution.}
  \vspace{-2mm}
  \label{fig:method_arch}
\end{figure}

\section{Methodology}

\subsection{Motivation}\label{sec:method_mot}
Given long-sequence, high-resolution inputs, most generalizable 3DGS methods patchify images into tokens and stack many layers to model token interactions for the prediction of depth and appearance attributes~\cite{ziwen2025long}. 
To meet high-fidelity NVS requirements, patches must be small to preserve detail, which imposes a heavy computational burden, since the cost of token interactions grows quadratically~\cite{vaswani2017attention} or linearly~\cite{gu2024mamba} with token length. 
This high computational demand motivates us to explore redundancies introduced by simplistic layer stacking, which not only increases computational overhead but also escalates parameter requirements. Therefore, this work aims to reduce such redundancy through strategic computation reallocation, thus improving parameter efficiency and enabling compact models to achieve strong performance.

\begin{figure}[t]
  \centering
  \includegraphics[width=1.0\linewidth]{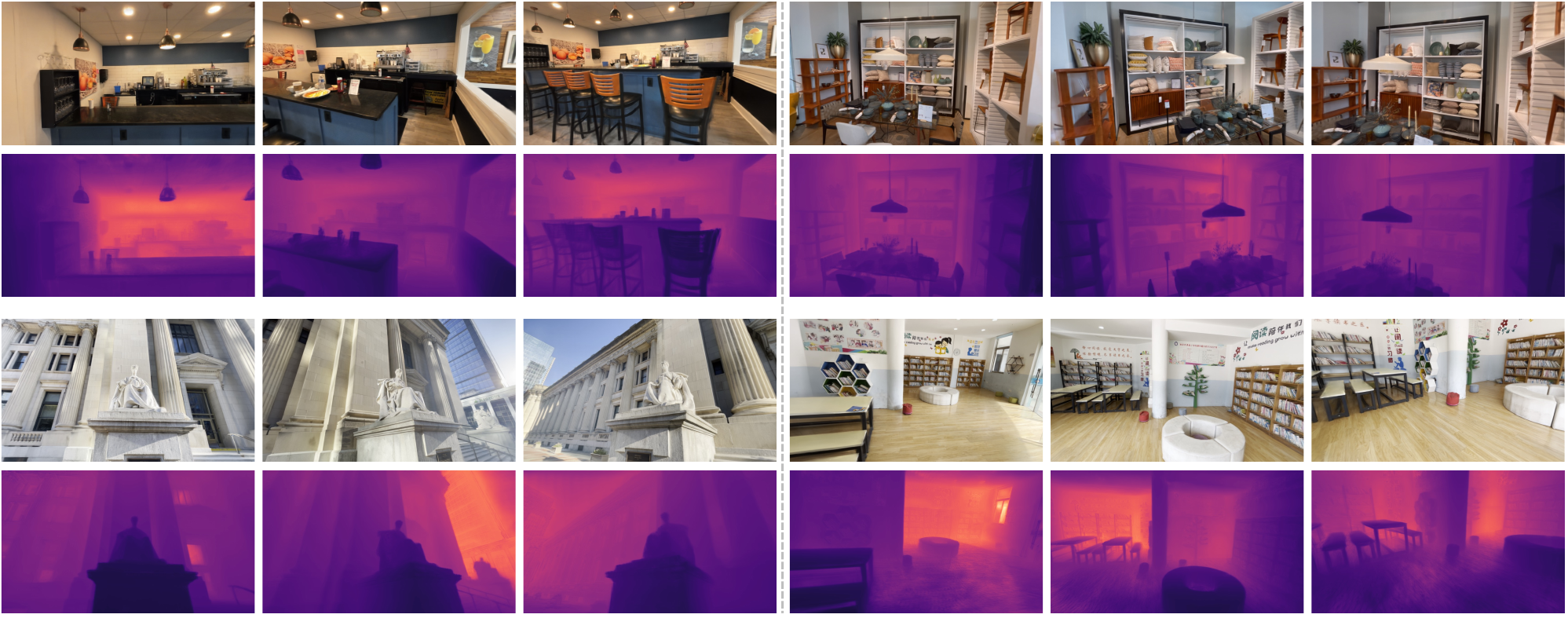}\vspace{-1mm}
  \caption{Rendering and depth prediction results of test views from AsySplat. The model demonstrates its capability to reconstruct the overall correct geometry of the scenes, with the coarse-grained tokens as input for the geometry branch. }
  \vspace{-2mm}
  \label{fig:depth_vis}
\end{figure}

Our computation allocation is guided by two observations in generalizable 3DGS: 
\textbf{(i)} 
High-precision geometry is not a must for high-fidelity NVS. Studies on surface reconstruction~\cite{guedon2024sugar,huang20242d} show that 3DGS can render well despite imperfect geometry, as shown in Fig.~\ref{fig:teaser}, largely due to alpha compositing of numerous Gaussians. This is further supported by our experiments, showing that our model achieves rendering quality comparable to state-of-the-art methods, even with geometry (Fig.~\ref{fig:depth_vis}) output from coarse-grained tokens. 
\textbf{(ii)} High-precision appearance attributes are necessary, while they are generally easier to model than geometry. Intuitively, once multi-view depth is established, the network only needs to map from 2D pixel-aligned cues to appearance attributes~\cite{yu2025wonderworld}, without inter-frame interaction. 
These insights inspire us to reallocate computation by decoupling geometry and appearance into two dedicated branches, each aligned with its precision needs and learning difficulty. 
This design leads to the asymmetric architecture depicted in Fig.~\ref{fig:method_arch}, with asymmetry manifested in input granularity, computation allocation, and parameter distribution.

\subsection{Asymmetric Gaussian Splatting Model}\label{sec:method_asym}
Given $V$ input images $\{\mathbf{I}_i\}_{i=1}^V$($\mathbf{I}_i \in \mathbb{R}^{H \times W \times 3}$) together with their camera intrinsics and extrinsics, our goal is to predict $N$ 3D Gaussians~\cite{kerbl20233d} to represent the entire scene:
$\{(\boldsymbol{\mu}_n, \alpha_n, \mathbf{s}_n, \mathbf{q}_n, \mathbf{SH}_n)\}_{n=1}^N$, corresponding to each Gaussian’s center, opacity, scale, quaternion, and spherical harmonics, respectively. 
The asymmetry in AsySplat starts from the tokenization stage. 
We first concatenate the images with their associated Plücker rays, and then tokenize the inputs at two different granularities. For the geometry branch, inputs are tokenized with a large patch size $p_c$, yielding coarse-grained tokens $\mathbf{C}^{0} \in \mathbb{R}^{(VHW/p_c^2) \times d_c}$, following observation \textbf{(i)} in Sec.~\ref{sec:method_mot}. In contrast, the appearance branch uses a smaller patch size $p_f$, where $p_f < p_c$, to preserve details for high-fidelity rendering, producing fine-grained tokens $\mathbf{F}^{0} \in \mathbb{R}^{V \times (HW/p_f^2) \times d_f}$. These two token sets are fed into the geometry and appearance branches, respectively. The subsequent asymmetric processing and fusion of these tokens are detailed below.

\noindent \textbf{Geometry Branch.} The geometry branch targets multi-view consistent depth recovery. Achieving this requires exhaustive matching across different views to establish correspondences. Therefore, the branch processes the entire sequence of coarse-grained tokens of length $VHW/p_c^2$. To balance accuracy and efficiency, the branch is structured into three hierarchical stages, with the token dimension progressively increasing through $\{d_s\}_{s=1}^3$. Within each stage, we adopt a hybrid architecture~\cite{hatamizadeh2025mambavision,ziwen2025long} that combines Mamba~\cite{gu2024mamba} with attention~\cite{vaswani2017attention} layers, enabling rapid acquisition of long-range context via efficient token interactions. Given the output $\mathbf{C}^{s-1}$ from the preceding stage $s-1$ ($s\in\{1,2,3\}$), stage $s$ performs the following operations: 
\begin{equation}\label{eq:geom}
\begin{split}
\mathbf{B}_0^{s}=\mathbf{C}^{s-1}, \mathbf{B}_{l}^{s}=\operatorname{Mamba}_{l}(\mathbf{B}_{l-1}^{s}), l=1,...,N_b,\\
\mathbf{T}_{0}^{s}=\mathbf{B}_{N_b}^{s}, \mathbf{T}_{l}^{s}=\operatorname{SparseAttn}_{l}(\mathbf{T}_{l-1}^{s}), l=1,...,N_a,\\
\end{split}
\end{equation}
where $N_b$ and $N_a$ denote the numbers of Mamba and attention layers per stage. 
Both $\mathbf{B}_l^{s}$ and $\mathbf{T}_l^{s}$ have the shape $(VHW/p_c^2) \times d_s$. 
We adopt bidirectional Mamba~\cite{zhu2024vision}, which performs forward and backward scans to provide each token with full context, and we employ a sparse attention module (Sec.~\ref{sec:method_spa}) for computational efficiency. 

In summary, the geometry branch stacks layers within three hierarchical stages to recover scene geometry. At the end of each stage, we introduce a bilateral connection module that exchanges information with the appearance branch. 
Given the intrinsic difficulty of multi-view geometric reasoning, we allocate the majority of AsySplat’s parameters to this branch.
It is important to note that observation \textbf{(i)} in Sec. \ref{sec:method_mot}, stating that `high-precision geometry is not a must', does not contradict the design of allocating more parameters to the geometry branch. This is because, even for coarse depth prediction (output from coarse-grained tokens), the network still needs to implicitly perform the challenging cross-view matching and multi-view stereo, which demands substantial network capacity. We validate the rationality of this design in our ablation studies.

\noindent \textbf{Appearance Branch.} In line with observation \textbf{(ii)} in Sec.~\ref{sec:method_mot}, inter-frame interaction is unnecessary for estimating appearance attributes. Consequently, we restrict token interactions to be intra-frame, so the effective sequence length during computation is bounded by the per-frame token count rather than the total across $V$ frames. Although the fine-grained token sequence $\mathbf{F}^{0}$ fed into the appearance branch is much longer than that of the geometry branch, its computational cost does not scale significantly with the number of views. This enables high-fidelity appearance modeling without incurring prohibitive multi-view overhead.
Mirroring the geometry pathway, the appearance branch is also organized into three stages. Each stage stacks several Mamba layers to strengthen per-frame appearance modeling. Given the output $\mathbf{F}^{s-1}$ from the preceding stage, stage $s$ applies the following operations: 
\begin{equation}\label{eq:app}
\mathbf{M}_0^{s}=\mathbf{F}^{s-1}, \mathbf{M}_{l}^s=\operatorname{MambaVision}_{l}(\mathbf{M}_{l-1}^s), l=1,...,N_v, 
\end{equation}
where $N_v$ denotes the number of layers per stage. $\mathbf{M}_l^s$ maintains shape of $V\times (HW/p_f^2) \times d_f$, and all operations in Eq.~\eqref{eq:app} are performed intra-frame. Unlike the hierarchical widths of the geometry branch, the appearance width $d_f$ is kept constant across stages. We adopt MambaVision~\cite{hatamizadeh2025mambavision} in this branch, which replaces causal convolutions with standard ones and adds a non–selective-scan branch. This design enhances global modeling capacity while being more efficient than bidirectional Mamba.

At the end of each stage, the fine-grained tokens exchange information with the coarse-grained tokens via a bilateral connection module. Consistent with observation \textbf{(ii)} in Sec.~\ref{sec:method_mot} that appearance attributes are generally easier to model, we allocate substantially fewer parameters to this branch (e.g., about 10$\%$ of the total), which shares design similarities with SlowFast~\cite{feichtenhofer2019slowfast} networks.

\noindent \textbf{Bilateral Connection Module.} The two branches communicate via bilateral connections. Coarse-grained tokens absorb high-frequency cues from the fine-grained tokens, while the latter receive geometric context to assist appearance attributes prediction, e.g., the scale of Gaussians is often correlated with its depth. As illustrated in Fig.~\ref{fig:method_arch}, the communication process can be formulated as: 
\begin{equation}\label{eq:bilat}
\begin{aligned}
&\Tilde{\mathbf{C}}^s=\operatorname{Reshape}(\mathbf{T}_{N_a}^s), \Tilde{\mathbf{F}}^s=\mathbf{M}_{N_v}^s,\\
&\mathbf{C}^s=\operatorname{Reshape}(\operatorname{CrossAttn}(\Tilde{\mathbf{C}}^s, \Tilde{\mathbf{F}}^s)),\\
&\mathbf{F}^s=\operatorname{SelfAttn}(\operatorname{CrossAttn}(\Tilde{\mathbf{F}}^s, \Tilde{\mathbf{C}}^s)), 
\end{aligned}
\end{equation}
where $\mathbf{T}_{N_a}^s$ and $\mathbf{M}_{N_v}^s$ are outputs from Eq.~\eqref{eq:geom} and Eq.~\eqref{eq:app}, respectively. The two 
$\operatorname{Reshape}$ operations convert tensors between shapes of $(VHW/p_c^2)\times d_s$ and $V\times (HW/p_c^2)\times d_s$, and vice versa. All computations in Eq.~\eqref{eq:bilat} are performed in an intra-frame manner.

\noindent \textbf{Merging and Gaussian Prediction.} To combine the complementary information from the two branches, we merge the two token sets to predict the final Gaussians. After reshaping them into 2D token maps of $\mathbf{C}^{\rm{out}}$ and $\mathbf{F}^{\rm{out}}$, we fuse them with a strided convolution to get an output map: 
\begin{equation}
    \mathbf{O}^{\rm{out}} = \mathbf{C}^{\rm{out}} + \operatorname{Conv}_{3\times3}(\mathbf{F}^{\rm{out}}, \operatorname{stride}=p_c/p_f), 
\end{equation}
whose shape is ${V \times (H/p_c) \times (W/p_c) \times d_3}$. Each pixel is mapped to $p_f^2$ Gaussians via a linear decoder. Rearranging this output yields a tensor of shape $V\times (Hp_f/p_c)\times (Wp_f/p_c)\times z$, $z$ denotes the number of parameters per Gaussian.
This tensor can be used to perform pixel-aligned predictions on images downsampled by a factor of $p_c/p_f$, resulting in a total of $VHW(p_f/p_c)^2$ Gaussians. For a configuration where $p_c > p_f$, the Gaussians are drastically reduced.

The asymmetry between the two branches is manifested in three aspects: input granularity, computational allocation, and parameter count. Our approach reduces computational redundancy by performing inter-frame interactions on coarse-grained tokens while confining fine-grained tokens to intra-frame processing. This contrasts with previous work~\cite{ziwen2025long}, which entangles fine-grained tokens in costly inter-frame computation. This more rational computational allocation promotes specialization, where parameters are dedicated to specific tasks, thus enhancing parameter efficiency.


\begin{figure}[t]
	\centering
	\includegraphics[width=0.98\linewidth]{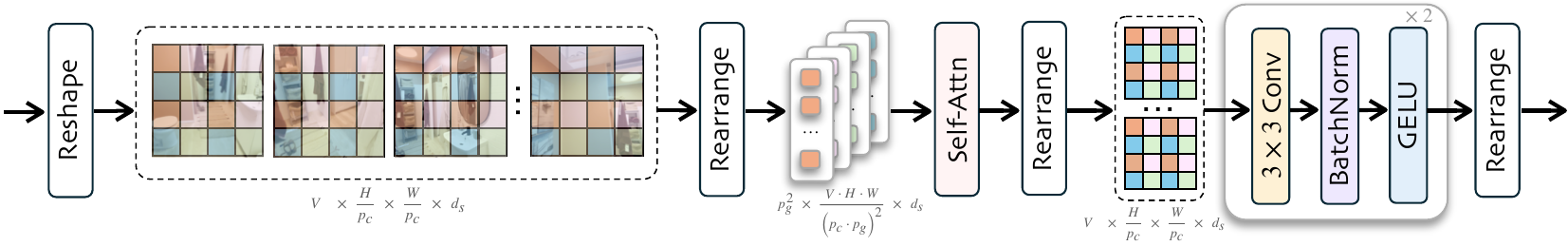}\vspace{-3mm}
	\caption{Illustration of the sparse attention module. The input token sequence is reshaped into 2D token maps, which are processed by two convolutional blocks to preserve contextual completeness. The refined token maps are partitioned with a uniform grid of size $p_g \times p_g$, which are rearranged into a sequence of reduced length for self-attention. $p_g$ is set to 2 for clarity in this illustration. }
	\vspace{-2mm}
	\label{fig:method_sp}
\end{figure}

\subsection{Sparse Attention Module}\label{sec:method_spa}
The proposed asymmetric structure reduces computational redundancy by allocating resources more judiciously. The saved computation can then be reinvested to enhance the model's capability. However, as noted by LongLRM~\cite{ziwen2025long}, excessive reliance on Mamba severely degrades performance, since state-space models are less effective for the dense matching required in multi-view depth estimation. Therefore, we retain more attention layers in each stage of our geometry branch.
Although the computational increase from these additional attention layers is acceptable, it leads to a substantial increase in training time. This is primarily due to the quadratic complexity with respect to the length of the coarse-grained token sequence: $\mathcal{O}((VHW/p_c^2)^2)$.

\begin{wraptable}{r}{0.5\textwidth}
	\vspace{-10pt}
	\caption{Training overhead across stages. $^*$ replaces sparse attention with full attention.}\vspace{-5pt}
	\centering
	\resizebox{1.\linewidth}{!}{
		\begin{tabular}{clccccc}
			\toprule[1pt]
			Image                    & \multicolumn{1}{l}{\multirow{2}{*}{Method}} & Batch & Train & \multirow{2}{*}{\#Param} & Iteration  & GPU  \\
			Size                     &                         & Size       & Step  &                          & Time  & Memory \\ \midrule[0.5pt]
			\multirow{2}{*}{256$\times$256} & LongLRM & \multirow{2}{*}{8} & 60k & 142M & 2.7s & 27G \\
			& AsySplat (Ours) & & 60k & \textbf{98M} & \textbf{1.9s} & \textbf{25G} \\ \midrule[0.5pt]
			\multirow{2}{*}{512$\times$512} & LongLRM & \multirow{2}{*}{2} & 10k & 142M & 2.9s & 27G \\
			& AsySplat (Ours) & & 10k & \textbf{98M} & \textbf{2.2s} & \textbf{25G} \\ \midrule[0.5pt]
			& LongLRM & \multirow{3}{*}{2} & 10k & 142M & 8.5s & 52G \\
			540$\times$960 & AsySplat (Ours) & & 10k & 98M & \textbf{6.0s} & \textbf{48G} \\
			& AsySplat$^*$ & & 10k & \textbf{95M} & 10.5s & \textbf{48G} \\
			\bottomrule[1pt]
		\end{tabular}
	}\vspace{-4mm}
	\label{tab:train_overhead}
\end{wraptable}

Inspired by MaxViT~\cite{tu2022maxvit}, which introduces grid attention to reduce computational overhead in single-frame processing, we extend this idea to multi-view inputs by developing a sparse attention module for inter-frame token interaction. 
As depicted in Fig.~\ref{fig:method_sp}, the input token sequence is first reshaped into 2D token maps. We then apply convolutional blocks to each frame's token map independently to aggregate local information. This step is crucial for preserving rich contextual details for every token. The refined token maps are partitioned using a uniform grid of size $p_g\times p_g$, which are subsequently rearranged into a structured sequence of shape $p_g^2 \times (VHW/(p_gp_c)^2) \times d_s$, on which self-attention is performed. Finally, the self-attention output is rearranged back to the original input token sequence shape. 
Experiments demonstrate that applying self-attention on this reduced-length sequence significantly shortens training time, while the convolutional refinement is also essential for context integrity and thus critical to performance.

\section{Experiments}

\subsection{Datasets}
DL3DV~\cite{ling2024dl3dv} is a large-scale dataset with over 10k training scenes and 140 scenes for benchmarking. We train AsySplat on the DL3DV training split and report most results on its benchmark split. To evaluate the zero-shot generalization capability of the model, we test on four datasets, including three widely adopted 3DGS benchmarks, Tanks\&Temples~\cite{knapitsch2017tanks}, MipNeRF-360~\cite{barron2022mip}, and Deep Blending~\cite{hedman2018deep}, as well as the synthetic Replica dataset~\cite{straub2019replica}.

\begin{table}[t]
\caption{Comparison with optimization-based 3DGS methods (the first block) and feed-forward LongLRM. `\#Iter' is the number of post-optimization iterations. This refinement process starts from the model's predicted Gaussians. `Time' is the total inference runtime. Image resolution is 540$\times$960. }\vspace{-0pt}\label{tab:sota}
\resizebox{1.\linewidth}{!}{
\centering
\begin{tabular}{lccccccccccccccccc}
\toprule[1pt]
\multirow{2}{*}{Method}         & \multirow{2}{*}{\#Iter} & \multirow{2}{*}{Time$\downarrow$} & \multicolumn{3}{c}{DL3DV-140~\cite{ling2024dl3dv}} & \multicolumn{3}{c}{Tanks\&Temples~\cite{knapitsch2017tanks}}&\multicolumn{3}{c}{MipNeRF-360~\cite{barron2022mip}}&\multicolumn{3}{c}{Deep Blending~\cite{hedman2018deep}} \\
                                &                       &                       & PSNR$\uparrow$     & SSIM$\uparrow$     & LPIPS$\downarrow$   & PSNR$\uparrow$       & SSIM$\uparrow$      & LPIPS$\downarrow$   &
                                PSNR$\uparrow$       & SSIM$\uparrow$      & LPIPS$\downarrow$   & PSNR$\uparrow$       & SSIM$\uparrow$      & LPIPS$\downarrow$ \\ \midrule[0.5pt]
3DGS~\cite{kerbl20233d}         & 30k                   & 13min                 & 23.60    & 0.779    & \cellcolor{orange!25}0.213   & 18.10      & \cellcolor{orange!25}0.688     & \cellcolor{red!25}\textbf{0.269} & 21.25 & 0.617 & \cellcolor{orange!25}0.255 & 21.00 & 0.713 & \cellcolor{orange!25}0.396   \\
Scaffold-GS~\cite{lu2024scaffold}  & 30k                   & 16min                 & 24.77    & 0.805    & \cellcolor{red!25}\textbf{0.205}   & 18.41      & \cellcolor{red!25}\textbf{0.691}     & \cellcolor{orange!25}0.290   & 21.53 & 0.630 & \cellcolor{red!25}\textbf{0.254} & 21.73 & \cellcolor{yellow!25}0.739 & \cellcolor{red!25}\textbf{0.361}  \\ \midrule[0.5pt]
                                & -                     & \textless{}1s                   & 24.10    & 0.783    & 0.254   & 18.38      & 0.601     & 0.363    & 21.54      & 0.529     & 0.417 & 20.91    & 0.701    & 0.464  \\
LongLRM~\cite{ziwen2025long}                         & 3                     & 15s                   & 24.99    & 0.809    & 0.243   & 18.69      & 0.623     & 0.360   & 22.49      & 0.601     & 0.388 & 21.38    & 0.715    & 0.456     \\
                                & 10                    & 50s                   & \cellcolor{orange!25}25.60    & \cellcolor{orange!25}0.826    & 0.233   & 18.90      & 0.642     & 0.350   & 23.13      & 0.632     & 0.373& 21.84    & 0.725    & 0.450      \\ \midrule[0.5pt]
\multirow{5}{*}{AsySplat (Ours)} & -                     & \textless{}1s       & 24.00    & 0.785    & 0.255   & 18.51      & 0.627     & 0.351 & 22.20      & 0.597     & 0.376   & 21.58    & 0.721    & 0.450     \\
                                & 3                     & 6s                    & 24.67    & 0.803    & 0.248   & 18.76      & 0.641     & 0.349 & 22.81      & 0.630     & 0.363  & 21.96    & 0.728    & 0.446      \\
                                & 10                    & 18s                   & 25.20    & \cellcolor{yellow!25}0.817    & 0.241   & \cellcolor{yellow!25}18.96      & 0.654     & 0.342   & \cellcolor{yellow!25}23.34      & \cellcolor{yellow!25}0.653     & 0.352 & \cellcolor{yellow!25}22.39    & 0.736    & 0.440          \\
                                & 20                    & 36s                   & \cellcolor{yellow!25}25.55    & \cellcolor{orange!25}0.826    & 0.231   & \cellcolor{orange!25}19.08      & 0.664     & 0.335   & \cellcolor{orange!25}23.71      & \cellcolor{orange!25}0.669     & 0.341  & \cellcolor{orange!25}22.77    & \cellcolor{orange!25}0.743    & 0.434    \\
                                & 28                    & 50s                   & \cellcolor{red!25}\textbf{25.72}    & \cellcolor{red!25}\textbf{0.830}    & \cellcolor{yellow!25}0.226   & \cellcolor{red!25}\textbf{19.14}      & \cellcolor{yellow!25}0.669     & \cellcolor{yellow!25}0.330   & \cellcolor{red!25}\textbf{23.89}      & \cellcolor{red!25}\textbf{0.678}     & \cellcolor{yellow!25}0.333  & \cellcolor{red!25}\textbf{22.95}    & \cellcolor{red!25}\textbf{0.746}    & \cellcolor{yellow!25}0.430    \\ 
\bottomrule[1pt]
\end{tabular}
}\vspace{-5mm}
\end{table}

\subsection{Implementation Details}
This work focuses on generalizable 3DGS for long-sequence inputs. Unless otherwise specified, all experiments use 32 input views and are evaluated at 540$\times$960.

\noindent \textbf{Architecture. }The patch sizes for tokenization, i.e., $p_c$ and $p_f$ are set to 16 and 8, respectively. The initial embedding dimensions $d_c$ and $d_f$ are set to 192 and 256, respectively. In the geometry branch, the token dimensions follow a hierarchical design across three stages. We set $\{d_s\}_{s=1}^3$ to $\{192, 512, 896\}$. In each stage, $N_b$ and $N_a$ are both set to 4. The expansion ratios in the sparse attention layers are set to 1, 2, and 3 for the three successive stages, respectively. In the appearance branch, we set $N_v$ to 2 and keep $d_f$ fixed at 256 across stages. For the sparse attention layers, we set the uniform grid size $p_g$ to 2.

\noindent \textbf{Training. }We adopt a progressive training strategy with increasing resolutions: 256$\times$256, 512$\times$512, and 540$\times$960. The batch size for each resolution stage is set to 256, 64, and 64. The training lasts for 60k, 10k, and 10k iterations. To improve training efficiency, we employ FlashAttention-2~\cite{dao2023flashattention}, gradient checkpointing~\cite{chen2016training}, and mixed-precision training with BFloat16. We utilize deferred backpropagation~\cite{zhang2022arf} during 3DGS rendering to reduce GPU memory consumption. For more training details, please refer to the supplementary materials. 

\begin{figure}[t]
  \centering
  \includegraphics[width=1.0\linewidth]{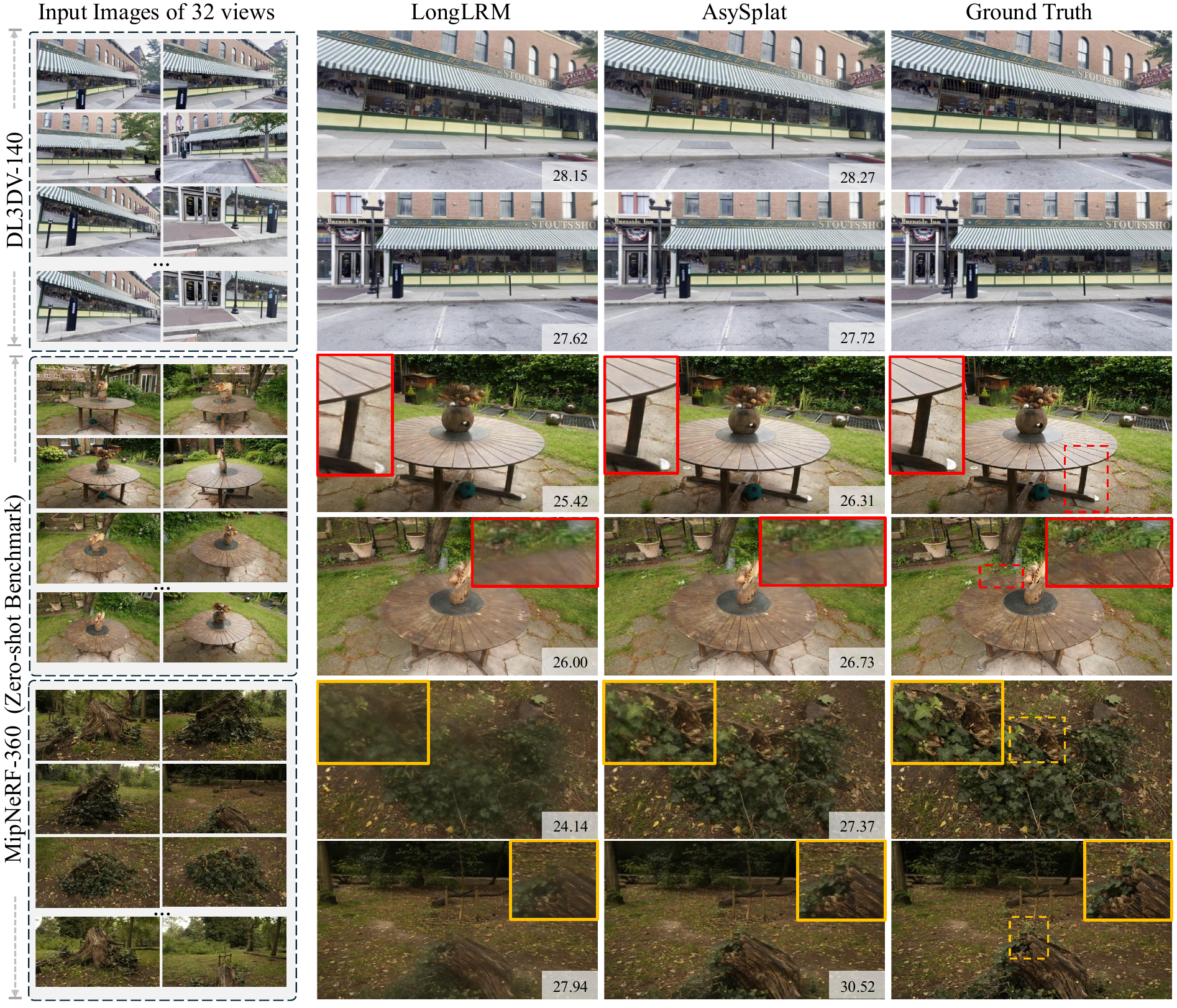}\vspace{-3mm}
  \caption{Qualitative comparison with LongLRM. Given 32 input views at a resolution of 540$\times$960, both models achieve comparable quality on the DL3DV-140 benchmark (the first scene), while AsySplat demonstrates superior performance on zero-shot benchmarks such as MipNeRF-360 (remaining scenes). Please refer to the demo for 
  	more visualization results. }
  \label{fig:rgb_vis}\vspace{-2mm}
\end{figure}

\subsection{Comparison with Current Methods}

\noindent \textbf{Comparison against optimization-based methods.} We compare AsySplat with optimization-based 3DGS methods~\cite{kerbl20233d,lu2024scaffold} on four datasets. On DL3DV-140, AsySplat achieves nearly 800$\times$ faster inference than 3DGS~\cite{kerbl20233d} (\textless{}1s vs. 13min) with higher PSNR (24.0~vs.~23.6). With only 10 post-optimization iterations (completed in 18s), PSNR further improves to 25.2, surpassing 24.8 of Scaffold-GS~\cite{lu2024scaffold}. Similar PSNR advantages are observed on other three datasets. On these datasets, AsySplat trails in LPIPS, which is common among generalizable 3DGS methods due to the lack of error-based densification. Overall, AsySplat matches optimization-based performance with significantly lower runtime.

\noindent \textbf{Comparison against generalizable 3DGS methods.} We compare our AsySplat with the state-of-the-art method LongLRM in the following. 

\begin{wraptable}{r}{0.5\textwidth}
	\vspace{-25pt}
	\caption{Comparison on synthetic Replica. Image resolution is at 540$\times$744.  No pruning is applied.}\vspace{-5pt}
	\centering
	\resizebox{1.\linewidth}{!}{
		\begin{tabular}{lcccccc}
			\toprule[1.2pt]
			Method & \#Iter & \#Param & \#GS & PSNR$\uparrow$ & SSIM$\uparrow$ & LPIPS$\downarrow$ \\ \midrule[0.5pt]
			LongLRM & 0 & 142M & 12M & 25.89 & 0.866 & 0.204 \\
			AsySplat (Ours) & 0 & 98M & 3M & \cellcolor{orange!40}\textbf{27.46} & \cellcolor{orange!40}\textbf{0.891} & \cellcolor{orange!40}\textbf{0.174} \\ \midrule[0.5pt]
			LongLRM & 10 & 142M & 12M & 28.37 & 0.897 & 0.179 \\
			AsySplat (Ours) & 10 & 98M & 3M &  \cellcolor{orange!40}\textbf{30.83} &  \cellcolor{orange!40}\textbf{0.921} &  \cellcolor{orange!40}\textbf{0.145} \\
			\bottomrule[1.2pt]
		\end{tabular}
	}\vspace{-3mm}
	\label{tab:sota_replica}
\end{wraptable}

\noindent \textbf{(i)} \emph{AsySplat achieves performance comparable to LongLRM.} As shown in Tab.~\ref{tab:sota}, with a feed-forward time less than 1s, both models achieve similar performance on DL3DV-140. This is confirmed by the similar rendering quality illustrated in Fig.~\ref{fig:rgb_vis}. 
When the number of post-optimization iterations is fixed, LongLRM yields better results. However, since AsySplat predicts fewer Gaussians (Tab.~\ref{tab:test}), it can afford more iterations under a similar time budget. In the following, for clarity, we use a subscript to denote the number of post-optimization iterations with the runtime in the bracket. Tab.~\ref{tab:sota} shows that, under comparable or lower total runtime, AsySplat remains competitive: AsySplat$_{10}$ (18 s) surpasses LongLRM$_{3}$ (15 s), AsySplat$_{20}$ (36 s) matches LongLRM$_{10}$ (50 s), and  AsySplat$_{28}$ (50 s) outperforms LongLRM$_{10}$ (50 s) under same time budget.

\begin{table*}[!t]
	\centering
	
	\begin{minipage}[t]{0.50\textwidth}
		\caption{Inference overhead. The `+post-optim' config includes a post-optimization of 10 iterations, initialized with the model's predicted Gaussians. }\label{tab:test}\vspace{-2mm}
		\centering
		\resizebox{1.0\linewidth}{!}{
			\begin{tabular}{clcccccc}
				\toprule[1.pt]
				Image                    & \multirow{2}{*}{Method} & \multirow{2}{*}{\#Param} & \multirow{2}{*}{Time}         & \multirow{2}{*}{TFLOPs} & GPU    & \multirow{2}{*}{\#GS}      \\
				Size                     &                         &                          &               &                         & Memory &                            \\ \midrule[0.5pt]
				\multirow{2}{*}{540$\times$960}                  & LongLRM                 & \multirow{2}{*}{142M}    & {\textless{}1s} & 20.2                    & 26G    & \multirow{2}{*}{8M} \\
				& +post-optim      &                          &  50s                  & -                       & 31G    &                            \\ \midrule[0.5pt]
				\multirow{2}{*}{540$\times$960} & AsySplat (Ours)                 & \multirow{2}{*}{{\textbf{98M}}}     & {\textless{}1s} & {\textbf{16.2}}                    & \textbf{22G}    & \multirow{2}{*}{{\textbf{4M}}} \\
				& +post-optim      &                          &  \textbf{18s}                & -                       & \textbf{27G}    &                            \\ 
				\bottomrule[1.2pt]
			\end{tabular}\vspace{-3mm}
		}
	\end{minipage}
	\hfill
	\begin{minipage}[t]{0.45\textwidth}
		\caption{Comparison with generalizable 3DGS methods under a sparse-view setting of two input views on RealEstate10K~\cite{zhou2018stereo}. }\label{tab:sparse}\vspace{-2mm}
		\centering
		\resizebox{0.94\linewidth}{!}{
			\begin{tabular}{lcccc}
				\toprule[1.pt] 
				Method         & $\#$Param & PSNR$\uparrow$  & SSIM$\uparrow$  & LPIPS$\downarrow$ \\ \midrule[0.5pt]
				pixelSplat~\cite{charatan2024pixelsplat}     & 125M    & 25.89 & 0.858 & 0.142 \\
				MVSplat~\cite{chen2024mvsplat}               & 12M     & 26.39 & 0.869 & 0.128 \\
				GS-LRM~\cite{zhang2025gs}                    & 300M    & 28.10 & \cellcolor{orange!40}\textbf{0.892} & 0.114 \\
				DepthSplat~\cite{xu2025depthsplat}           & 354M    & 27.47 & 0.889 & 0.114 \\
				AsySplat (Ours)                               & 98M     & \cellcolor{orange!40}\textbf{28.27} & \cellcolor{orange!40}\textbf{0.892} & \cellcolor{orange!40}\textbf{0.105}      \\
				\bottomrule[1.pt]
			\end{tabular}
		}
	\end{minipage}\vspace{-3mm}
	
\end{table*}

\noindent \textbf{(ii)} \emph{AsySplat exhibits superior zero-shot capabilities.} As shown in Tab.~\ref{tab:sota}, on the Tanks\&Temples dataset, our model achieves better performance than LongLRM in a single feed-forward pass, with higher scores across all metrics. Moreover, with the same number of post-optimization iterations, AsySplat not only delivers higher quality but also requires less total runtime.
On Deep Blending, AsySplat maintains superior performance under all settings, whether under the same post-optimization iterations or a similar time budget. Notably, AsySplat$_{20}$ outperforms LongLRM$_{10}$ by nearly 1.0 PSNR. A similar advantage is observed on MipNeRF-360, where AsySplat surpasses LongLRM by nearly 0.7 PSNR in a single feed-forward pass. Notably, on the synthetic Replica, our method shows a clear advantage over LongLRM both with and without post-optimization. In particular, with 10 post-optimization iterations, it achieves an over 2.5 PSNR advantage over LongLRM. A qualitative comparison is shown in Fig.~\ref{fig:rgb_vis}.

\noindent \textbf{(iii)} \emph{AsySplat exhibits lower training overhead.} AsySplat and LongLRM are both trained using a progressive strategy with increasing image resolutions. We measure the training overhead at each stage, as summarized in Tab.~\ref{tab:train_overhead}. Owing to a more efficient allocation of computational resources and a more compact model design, AsySplat consistently reduces both iteration time and GPU memory usage throughout training. At a resolution of 540$\times$960, for example, AsySplat decreases iteration time by nearly 30$\%$ compared to LongLRM.

\noindent By accumulating the iteration time across all training stages, the total estimated training time for LongLRM is 76 hours, while AsySplat requires only 54 hours. It should be noted that the iteration time in Tab.~\ref{tab:train_overhead} does not include inter-node communication time, which scales with model parameters. As a result, in a real distributed training setup, the actual time reduction achieved by AsySplat would be even more pronounced.

\noindent \textbf{(iv)} \emph{AsySplat exhibits lower inference overhead.} Tab.~\ref{tab:test} shows that AsySplat achieves reductions in TFLOPs, GPU memory usage, and the number of Gaussians. These improvements stem from a more compact model design and the more judicious computation allocation. Furthermore, since AsySplat predicts only half the Gaussians, its post-optimization completes much faster. In summary, these factors collectively make AsySplat a more efficient and lightweight solution during inference.

\noindent \textbf{(v)} \emph{AsySplat exhibits strong sparse-view performance. }Although our designs primarily focus on reallocating computation for long-sequence inputs, and improve overall efficiency, our framework also achieves strong performance under sparse-view settings, such as with 2-view inputs on RealEstate10K~\cite{zhou2018stereo}. The results are presented in Tab.~\ref{tab:sparse}. Notably, AsySplat not only delivers strong performance, but also offers clear advantages in parameter efficiency.

\begin{table*}[!t]
\centering
\caption{Model analysis with 32 input views at 256$\times$256 resolution. All experiments are conducted on DL3DV-140. }\label{tab:abla}\vspace{-1mm}
\begin{subtable}[t]{0.50\textwidth}
\centering
\resizebox{1.0\linewidth}{!}{
\begin{tabular}{lcccc}
\toprule[1pt]
Method & \#Param & PSNR$\uparrow$  & SSIM$\uparrow$ & LPIPS$\downarrow$ \\
\midrule[0.5pt]
AsySplat & 98M & \cellcolor{orange!40}\textbf{19.24} & \cellcolor{orange!40}\textbf{0.53} & \cellcolor{orange!40}\textbf{0.49} \\
Pred. from fine   & 98M & 19.02 & 0.52 & 0.50 \\
Pred. from coarse & 95M & 18.54 & 0.47 & 0.54 \\
Cross-attn w/ fine-to-coarse   & {89M} & 19.10 & 0.52 & \cellcolor{orange!40}\textbf{0.49} \\
Cross-attn w/ coarse-to-fine   & {89M} & 17.33 & 0.44 & 0.54 \\
w/o MambaVision    & 94M & 19.06 & 0.51 & 0.51 \\
\bottomrule[1pt]
\end{tabular}}
\caption{{Ablations} on prediction sources for Gaussians (coarse-grained or fine-grained tokens), cross-attention links in the bilateral connection module, and MambaVision in the appearance branch. }
\label{tab:abl_arch}
\end{subtable}
\hfill
\begin{subtable}[t]{0.45\textwidth}
\centering
\resizebox{0.98\linewidth}{!}{
\begin{tabular}{lcccc}
\toprule[1pt]
Method & \#Param & PSNR$\uparrow$ & SSIM$\uparrow$ & LPIPS$\downarrow$ \\
\midrule[0.5pt]
AsySplat (4m+4s)     & 98M  & \cellcolor{orange!40}\textbf{19.24} & \cellcolor{orange!40}\textbf{0.53} & \cellcolor{orange!40}\textbf{0.49} \\
7m+1s               & 88M  & 18.74 & 0.50 & 0.52 \\
7m+1t               & {87M}  & 18.78 & 0.50 & 0.52 \\
4m+4t               & 95M  & 19.19 & 0.52 & 0.50 \\
8t                  & 105M & 19.16 & 0.52 & \cellcolor{orange!40}\textbf{0.49} \\
4m+4s (no conv)     & 95M  & 17.78 & 0.44 & 0.58 \\ 
\bottomrule[1pt]
\end{tabular}}
\caption{{Layer config per stage. }m and t/s represent bidirectional Mamba and full/sparse attention. `no conv' indicates removing convolution blocks in the sparse attention module.}\label{tab:abl_layer}
\end{subtable}
\vspace{4pt}
\centering
\begin{subtable}[t]{0.48\textwidth}
\vspace{0pt}
\centering
\resizebox{0.99\linewidth}{!}{%
\renewcommand{\arraystretch}{1.1}%
\begin{tabular}{lcccc}
\toprule[1.1pt]
Method & PSNR$\uparrow$ & $\delta<1.25 \uparrow$ & AbsRel$\downarrow$ & GFLOPs$\downarrow$ \\
\midrule[0.5pt]
Single branch & 19.11 & 0.32 & 0.69 & 6104 \\
Two branches & \cellcolor{orange!40}\textbf{19.24} & \cellcolor{orange!40}\textbf{0.49} & \cellcolor{orange!40}\textbf{0.44} & 5892 \\
Asym. branches (Ours) & \cellcolor{orange!40}\textbf{19.24} & 0.41 & 0.55 & \cellcolor{orange!40}\textbf{2033} \\
\bottomrule[1.1pt]
\end{tabular}
}
\caption{Validation of the asymmetric branch design. Input patch sizes: Single branch $p\!=\!8$, Two branches $(p_c,p_f)\!=\!(8,8)$, and Asymmetric branches $(p_c,p_f)\!=\!(16,8)$.}\label{tab:abl_design}
\end{subtable}
\hspace{12pt}
\begin{subtable}[t]{0.45\textwidth}
\vspace{0pt}
\centering
\resizebox{0.95\linewidth}{!}{
\centering
\begin{tabular}{lcccc}
\toprule[1pt]
Method & \#Param & PSNR$\uparrow$ & SSIM$\uparrow$ & LPIPS$\downarrow$ \\
\midrule[0.5pt]
AsySplat-30  & {98M}  & \cellcolor{orange!40}\textbf{19.24} & \cellcolor{orange!40}\textbf{0.53} & \cellcolor{orange!40}\textbf{0.49} \\
LongLRM-18       & 103M & 18.39 & 0.48 & 0.53 \\
LongLRM-24       & 99M  & 18.39 & 0.48 & 0.53 \\
LongLRM-24\# & 102M & 17.71 & 0.43 & 0.59 \\ 
\bottomrule[1pt]
\multicolumn{5}{c}{} \\
\multicolumn{5}{c}{} \\ 
\end{tabular}}\vspace{-14pt}
\caption{
Model comparison with similar parameters. The number indicates total layers. $\#$ denotes the use of hierarchical parameters across stages.
}\label{tab:abl_param}
\end{subtable}
\vspace{-7mm}
\end{table*}

\subsection{Model Analysis}
We present a thorough model analysis in Tab.~\ref{tab:abla}. Due to resource constraints, all experiments in this section are conducted on 32 input views at a resolution of 256$\times$256, and models are trained with a batch size of 24 for 10k iterations.

\noindent \textbf{Architecture ablations.} We ablate three of our architecture's designs in this part, i.e., the merging operation for Gaussian prediction, the bilateral connection module for cross-branch interactions, and the MambaVision for appearance modeling. As shown in Tab.~\ref{tab:abl_arch}, using only coarse-grained tokens from the geometry branch or only fine-grained tokens from the appearance branch for prediction degrades performance. 
Specifically, relying solely on coarse-grained tokens results in a noticeable PSNR drop of 0.7. Conversely, using only fine-grained tokens without geometric guidance also yields suboptimal results. 
This indicates that the outputs from the two branches are complementary for the prediction of Gaussians. We further evaluate the bilateral connections by removing one cross-attention link in Fig.~\ref{fig:method_arch}. Removing either cross-attention link also harms performance, confirming the contribution of each link in the bilateral design. 
Finally, the performance decline observed by removing MambaVision in the appearance branch confirms its contribution to appearance modeling. 

\noindent \textbf{Layer configuration for each stage.} AsySplat processes tokens of different granularities through three stages, each composed of 8 layers. We investigate the impact of different layer configurations on performance, with results shown in Tab.~\ref{tab:abl_layer}. Increasing the proportion of bidirectional Mamba layers (`7m+1s' or `7m+1t') leads to a performance drop, consistent with findings in LongLRM~\cite{ziwen2025long}. Replacing sparse attention with full attention (`4m+4t') yields similar accuracy but substantially increases training time, as shown in Tab.~\ref{tab:train_overhead}, where iteration time rises from 6s to 10.5s. Using only full attention (`8t') further slows training without bringing accuracy gains. As discussed in Sec.~\ref{sec:method_spa}, the convolution blocks in the sparse attention module are critical for context completeness, which is confirmed by the over 1.0 PSNR drop of the `4m+4s (no conv)' variant.

\noindent \textbf{Design principles verification.}
The asymmetric design is motivated by the two observations in Sec.~\ref{sec:method_mot}. We verify them in Tab.~\ref{tab:abl_design} with rendering and geometry metrics.
\textbf{(i)} Although `Two branches' achieves more accurate geometry than `Asymmetric branches', it does not improve NVS quality and requires higher FLOPs, supporting that \emph{coarse geometry is sufficient}.
\textbf{(ii)} Both `Two branches' and `Asymmetric branches' use intra-frame-only interaction in the appearance branch, whereas `Single branch' uses inter-frame interaction for both geometry and appearance. Their stronger performance suggests that \emph{fine appearance is easier to model and can rely on intra-frame-only interaction}.

\noindent \textbf{Models with similar parameters.} This work explores redundancy reduction through computational reallocation, enabling compact models to achieve strong performance. To demonstrate that designing an effective compact model is non-trivial, we compare AsySplat (98M, with a 24-layer geometry branch and a 6-layer appearance branch) against several parameter-reduced variants of LongLRM~\cite{ziwen2025long} (the official 142M, 24-layer model). These include a shallower 18-layer model, a 24-layer model with reduced feature dimensions, and a model using the same hierarchical parameters across three stages as AsySplat. Tab.~\ref{tab:abl_param} shows that AsySplat outperforms all these variants, confirming that its more rational computation allocation dedicates parameters to specific tasks more effectively, thereby achieving high performance with improved parameter efficiency.

\section{Conclusion}\vspace{-2pt}
This work studies computation redundancy in generalizable 3D Gaussian Splatting methods for long-sequence, high-resolution novel view synthesis. We propose an asymmetric architecture that decouples the modeling of geometry and appearance into two branches, and allocates the computation and parameters based on their respective precision needs and learning difficulty. This design improves parameter efficiency and thus enables smaller models to achieve strong performance. Extensive experiments demonstrate that our approach matches current generalizable models with markedly fewer parameters and lower overhead, while also exhibiting superior zero-shot performance.

	
\bibliographystyle{plainnat}
\bibliography{main}

\newpage
\clearpage
\makeatletter
\vbox{%
  \hsize\textwidth
  \linewidth\hsize
  \vskip 0.1in
  \@toptitlebar
  \centering
  {\LARGE\bf AsySplat: Efficient Asymmetric 3D Gaussian Splatting for Long-Sequence Scene Modeling\par}
  \@bottomtitlebar
  {\large\bf Supplementary Material\par}
  \vskip 0.1in
}
\makeatother
\renewcommand{\thefigure}{A\arabic{figure}}
\renewcommand{\thetable}{A\arabic{table}}
\renewcommand{\thesection}{\Alph{section}}
\renewcommand{\theHfigure}{supp.\arabic{figure}}
\renewcommand{\theHtable}{supp.\arabic{table}}
\renewcommand{\theHsection}{supp.\Alph{section}}
\setcounter{figure}{0}
\setcounter{table}{0}
\setcounter{section}{0}

\section{Implementation Details}

\noindent \textbf{Architecture.} All transformer blocks in AsySplat, including the sparse attention module in the geometry branch and the full attention layers in the bilateral connection module, apply RMSNorm~\cite{zhang2019root} after the query and key projections~\cite{touvron2023llama}. The geometry branch utilizes bidirectional Mamba~\cite{zhu2024vision} layers with a state dimension of 256, while the appearance branch employs MambaVision~\cite{hatamizadeh2025mambavision} layers with a state dimension of 16. For Gaussian attribute prediction, we set the degree of spherical harmonics to 3.

\noindent \textbf{Training.} We employ a progressive training strategy, increasing the resolution sequentially: 256$\times$256, 512$\times$512, and 540$\times$960. The model is trained using the AdamW optimizer~\cite{loshchilov2017decoupled} with a weight decay of 0.05 across all stages. The peak learning rates for the three stages are set to 4e-4, 4e-5, and 4e-5, respectively, and are decayed following a cosine schedule. Additionally, a linear warmup of the learning rate is applied for the first 2k iterations in the initial stage.

\noindent \textbf{Training overhead.} In the paper, for overhead comparisons with LongLRM~\cite{ziwen2025long} at 256$\times$256 and 512$\times$512 resolutions, we report overhead based on randomly initialized parameters, averaging the first 20 iterations due to the unavailability of LongLRM checkpoints for these resolutions. Generally speaking, the computational overhead of LongLRM increases more significantly than that of AsySplat as training progresses. This is because predicting 4$\times$ Gaussians leads to a steeper rise in GPU memory consumption and iteration time. For the overhead at 540$\times$960, we use the officially released LongLRM checkpoint. We compare it with our final model by continuing training for 20 iterations and report the average overhead.

\noindent \textbf{Inference overhead.} We compare the inference overhead of AsySplat with LongLRM at a resolution of 540$\times$960. The evaluation metrics reported in the main paper include inference time, TFLOPs, GPU memory usage, and the number of Gaussians. For inference time, we compute the average over the DL3DV-140 benchmark~\cite{ling2024dl3dv} after excluding the longest time as an outlier. For TFLOPs, we measure the computational cost of the network using the $\rm{thops}$ library. This measurement does not include the computation in the splatting. Note that at 540$\times$960, LongLRM initially predicts 4$\times$ as many Gaussians as our method, but then prunes them from 16M to 8M based on predicted opacities.

\noindent \textbf{Evaluation settings.} We follow LongLRM to determine the input and test splits for all evaluations, whether on the DL3DV-140 benchmark or other zero-shot datasets~\cite{knapitsch2017tanks,hedman2018deep,barron2022mip,straub2019replica}. For a given set of scene images, we uniformly sample every 8th image in the sequence as the test split. The input views are selected from the remaining images using K-means clustering based on camera positions and viewing directions, aiming to achieve optimal scene coverage. The number of clusters is set equal to the number of input views, and the cameras closest to the cluster centers are chosen as the input split. In the ablation experiments conducted on DL3DV-140 at 256$\times$256 resolution, we use the first 96 frames of each video sequence. Every 8th frame is uniformly selected as the test split, and 32 input views are uniformly sampled from the remaining frames. Regarding the comparisons on the Replica dataset, we select scenes following ~\cite{zhong2025taming}.

\noindent \textbf{Post optimization.} To ensure a fair comparison with LongLRM, we apply an identical post-optimization procedure using same hyperparameters. Specifically, we render the input views from the predicted Gaussians and compute the MSE loss between the rendered and ground truth images. The loss is then backpropagated to update the Gaussian parameters. The learning rates for the center and spherical harmonics are set to 5e-4 and 1e-3, respectively, while those for opacity, scale, and quaternion are set to 0.

\noindent \textbf{Training objective.} To supervise AsySplat, we render predicted Gaussians from $V_t$ target poses, obtaining $\{\hat{\mathbf{I}}_i\}_{i=1}^{V_t}$, which are compared against the corresponding ground truth images $\{\mathbf{I}_i\}_{i=1}^{V_t}$. Following prior work~\cite{zhang2025gs,ziwen2025long}, we use an MSE term and a perceptual loss~\cite{johnson2016perceptual,chen2017photographic}:
$
\mathcal{L}_{\rm{image}} = \frac{1}{V_t}\sum_{i=1}^{V_t} \operatorname{MSE}(\hat{\mathbf{I}}_i, \mathbf{I}_i) + \lambda_p\cdot\operatorname{Perceptual}(\hat{\mathbf{I}}_i, \mathbf{I}_i),
$
where $\lambda_p$ is a weighting factor. Following LongLRM~\cite{ziwen2025long}, we also adopt geometry supervision from monocular disparity estimates~\cite{yang2024depth} for training stability:
$
\mathcal{L}_{\rm{depth}} = \frac{1}{V_t}\sum_{i=1}^{V_t} \operatorname{SmoothL1}(\hat{\mathbf{D}}_i, \mathbf{D}_i),
$
where $\hat{\mathbf{D}}_i$ and $\mathbf{D}_i$ denote the normalized predicted and monocular-estimated disparity maps. Both maps are downsampled by a factor of $p_c/p_f$. Therefore, the overall objective is:
\begin{equation}\label{eq:loss_all_sup}
\mathcal{L} = \mathcal{L}_{\rm{image}} + \lambda_d\cdot \mathcal{L}_{\rm{depth}},
\end{equation}
where $\lambda_d$ balances the depth supervision. Note that although we incorporate monocular depth supervision, AsySplat still models coarse-grained geometry due to the coarse-grained input tokens. Moreover, the downsampled output $\hat{\mathbf{D}}_i$ provides an even coarser representation of the overall scene geometry, since the scene geometry is modeled by pixel-aligned prediction. The weighting factors $\lambda_p$ and $\lambda_d$ are set to 0.5 and 0.01. 

\section{More Comparisons}
\noindent \textbf{Comparison with pose-free methods.} Our method focuses on pose-given generalizable 3DGS modeling, we thus do not report comparisons with current pose-free generalizable 3DGS methods in the main paper. The setting mismatch can cause large performance gaps and maybe misleading. To demonstrate this, we show the comparisons with current state-of-the-art pose-free method AnySplat~\cite{jiang2025anysplat} in Tab.~\ref{tab:abla_anysplat}. The performance gap demonstrates that it is not fair to directly conduct comparisons. Moreover, our method has a different research focus with AnySplat. AnySplat fine-tunes a pretrained VGGT~\cite{wang2025vggt} into a 3DGS model, whereas we emphasize efficiency by training a parameter-efficient model from scratch with much lower training/inference overhead. Since our method prioritizes efficiency, whether using 32, 64, or even more input views, our approach maintains significantly lower memory and faster inference (e.g., 50$\times$ faster for 64 views) compared to AnySplat.

\begin{table}[t]
\centering
\caption{Comparison with AnySplat using the official inference code. }\label{tab:abla_anysplat}\vspace{-0mm}
\resizebox{0.7\linewidth}{!}{
\begin{tabular}{l|c|c|ccc|ccc}
\toprule[1pt]
         & Input     & Given          & \multicolumn{3}{c|}{DL3DV-140}  & Mem               & Param & Time  \\
         & Views      &Poses          & PSNR$\uparrow$ & SSIM$\uparrow$ & LPIPS$\downarrow$ & (G) & (M) & (s)   \\ \midrule[0.5pt]
AnySplat~\cite{jiang2025anysplat} & \multirow{2}{*}{32} & $\times$ & 13.91 & 0.392 & 0.609 & \textgreater{}32 & 1190 & 7.13 \\
AsySplat (Ours) & & \checkmark & \cellcolor{orange!40}\textbf{24.00} & \cellcolor{orange!40}\textbf{0.785} & \cellcolor{orange!40}\textbf{0.256} & \cellcolor{orange!40}\textbf{22} & \cellcolor{orange!40}\textbf{98} & \cellcolor{orange!40}\textbf{0.27} \\ \midrule[0.5pt]
AnySplat~\cite{jiang2025anysplat} & \multirow{2}{*}{64} & $\times$ & 13.98 & 0.405 & 0.613 & \textgreater{}56 & 1190 & 24.74 \\
AsySplat (Ours) & & \checkmark & \cellcolor{orange!40}\textbf{19.84} & \cellcolor{orange!40}\textbf{0.688} & \cellcolor{orange!40}\textbf{0.362} & \cellcolor{orange!40}\textbf{26} & \cellcolor{orange!40}\textbf{98} & \cellcolor{orange!40}\textbf{0.47} \\ \bottomrule[1pt]
\end{tabular}
}\vspace{-3mm}
\end{table}

\section{More Ablations}

\begin{table}[t]
\centering
\caption{Ablations on Mamba ratio in each stage. Each stage contains 8 layers, involving Mamba and attention layers. No sparse attention is used. Depth metrics are reported as indication of capbility of cross-view matching. }\label{tab:abla_mamba}\vspace{-0mm}
\resizebox{0.8\linewidth}{!}{
\begin{tabular}{cc|c|c|cc|cc|cc}
\toprule[1pt]
\multirow{2}{*}{\#Mamba} & \multirow{2}{*}{\#Attn} & \multirow{2}{*}{Abbrev.} & Param & \multicolumn{2}{c|}{DL3DV-140} & \multicolumn{2}{c|}{Mip-NeRF360} & \multicolumn{2}{c}{ScanNet++} \\
                         &                         &                          & (M)   & PSNR$\uparrow$ & SSIM$\uparrow$ & PSNR$\uparrow$ & SSIM$\uparrow$ & AbsRel$\downarrow$ & $\delta$\textless1.25$\uparrow$ \\ \midrule[0.5pt]
8 & 0 & 8m0a & 93.6 & 18.24 & 0.47 & 16.65 & 0.30 & 0.64 & 0.21 \\
7 & 1 & 7m1a & 96.6 & 18.80 & 0.49 & 16.91 & 0.29 & 0.57 & 0.35 \\
6 & 2 & 6m2a & 99.6 & 19.04 & 0.51 & 16.92 & 0.29 & 0.55 & 0.36 \\
4 & 4 & 4m4a & 95.0 & 19.19 & \cellcolor{orange!40}\textbf{0.52} & 17.06 & 0.30 & 0.44 & 0.39 \\
2 & 6 & 2m6a & 99.9 & \cellcolor{orange!40}\textbf{19.30} & \cellcolor{orange!40}\textbf{0.52} & \cellcolor{orange!40}\textbf{17.49} & \cellcolor{orange!40}\textbf{0.32} & \cellcolor{orange!40}\textbf{0.39} & \cellcolor{orange!40}\textbf{0.44} \\
0 & 8 & 0m8a & 105.0 & 19.16 & \cellcolor{orange!40}\textbf{0.52} & 17.42 & 0.29 & 0.43 & 0.40 \\ \bottomrule[1pt]
\end{tabular}
}\vspace{-3mm}
\end{table}

\begin{table}[t]
\centering
\caption{Ablations on parameter allocation between the geometry and appearance branches, with similar total parameters.}\label{tab:abla_ratio}\vspace{-0mm}
\resizebox{0.6\linewidth}{!}{
\begin{tabular}{cccccc}
\toprule[1pt]
Geometry & \#Param & \multicolumn{2}{c}{DL3DV-140} & \multicolumn{2}{c}{MipNeRF-360} \\
Branch Ratio    &    & PSNR$\uparrow$         & SSIM$\uparrow$        & PSNR$\uparrow$           & SSIM$\uparrow$          \\ \midrule[0.5pt]
90$\%$     & 98M & \cellcolor{orange!40}\textbf{19.24} & \cellcolor{orange!40}\textbf{0.53} & \cellcolor{orange!40}\textbf{17.52} & 0.30 \\
80$\%$     & 96M & 19.16 & 0.52 & 17.30 & \cellcolor{orange!40}\textbf{0.31} \\
70$\%$     & 98M & 19.20 & 0.52 & 16.34 & 0.27 \\
60$\%$     & 98M & 19.22 & \cellcolor{orange!40}\textbf{0.53} & 16.65 & 0.27 \\
50$\%$     & 96M & 18.97 & 0.51 & 16.50 & 0.27 \\ \bottomrule[1pt]
\end{tabular}
}\vspace{-3mm}
\end{table}

\noindent \textbf{Ablations on Mamba ratio.} In the main paper, we claim that `excessive reliance on Mamba degrades performance, since state-space models are less effective for the explicit matching required by multi-view depth estimation.' To substantiate this claim, we conduct experiments in this section by varying the Mamba ratio at each stage while keeping the overall parameter count similar. The results are presented in Tab.~\ref{tab:abla_mamba}. For novel view synthesis, we observe a sweet spot at 2m6a, beyond which increasing the Mamba ratio (from 2m6a to 8m0a) leads to a monotonic decline in NVS performance, consistent with our assertion that `\emph{excessive reliance} on Mamba degrades performance'. To further demonstrate that this degradation is linked to Mamba's ineffectiveness in explicit matching, we also report depth estimation metrics, as depth accuracy serves as a reliable indicator of matching quality. As the Mamba ratio increases beyond 2m6a, depth accuracy declines, following a trend similar to that of the NVS metrics. The deterioration in geometric accuracy strongly suggests that \emph{excessive reliance} on Mamba impairs the model's cross-view matching capability within the geometry branch.

\noindent \textbf{Parameter allocation ablations.} Our method adopts an asymmetric structure. In this part, we ablate the parameter ratio between the geometry branch and the appearance branch. We vary the proportions of the branches while keeping the total parameter count roughly unchanged. The experimental results are shown in Tab.~\ref{tab:abla_ratio}. The results indicate that allocating 90$\%$ of the parameters to the geometry branch achieves a decent performance, across benchmakrs.

\section{Discussion}
This work investigates computational redundancy in generalizable 3D Gaussian Splatting methods for long-sequence, high-resolution novel view synthesis. Although the proposed AsySplat achieves comparable or superior performance to state-of-the-art generalizable 3DGS approaches with substantially fewer parameters and lower overhead, its visual fidelity can still lag behind optimization-based methods, for example, yielding higher LPIPS. This limitation is common among generalizable 3DGS methods, which typically lack error-driven feedback for densification, a mechanism that is crucial for recovering fine details. Incorporating this capability could effectively leverage the parameters saved by AsySplat, and we leave this as future work.
\end{document}